%% file: main.tex
\documentclass[sigconf]{acmart}

\usepackage[utf8]{inputenc} 
\usepackage[T1]{fontenc}    
\usepackage{hyperref}       
\usepackage{url}            
\usepackage{booktabs}       
\usepackage{amsfonts}       
\usepackage{nicefrac}       
\usepackage{microtype}      
\usepackage{xcolor}         

\usepackage{bbm}
\usepackage{algorithm}
\usepackage{algpseudocode}
\usepackage{enumitem}
\usepackage{subcaption}

\input{math_commands}

\usepackage{amsmath}
\usepackage{mathtools}
\usepackage{amsthm}

\theoremstyle{plain}
\newtheorem{theorem}{Theorem}
\newtheorem{proposition}{Proposition}
\newtheorem{lemma}{Lemma}
\newtheorem{corollary}{Corollary}
\newtheorem{definition}{Definition}
\newtheorem{condition}{Condition} 
\theoremstyle{remark}

\newcommand{\indep}{\perp\!\!\!\!\perp} 
\usepackage{pifont}
\newcommand{\cmark}{\ding{51}}%
\newcommand{\xmark}{\ding{55}}%

\usepackage{subcaption}

\AtBeginDocument{%
  }

\setcopyright{acmlicensed}

\copyrightyear{2026}
\acmYear{2026}
\setcopyright{cc}
\setcctype{by}
\acmConference[KDD 2026] {Proceedings of the 32nd ACM SIGKDD Conference on Knowledge Discovery and Data Mining V.2}{August 9--13, 2026}{Jeju Island, Republic of Korea.}
\acmBooktitle{Proceedings of the 32nd ACM SIGKDD Conference on Knowledge Discovery and Data Mining V.2 (KDD 2026), August 9--13, 2026, Jeju Island, Republic of Korea}

\acmConference[KDD '26]{Proceedings of the 32nd ACM SIGKDD Conference on Knowledge Discovery and Data Mining}{August 9--13, 2026}{Jeju, Korea}

\acmISBN{979-8-4007-2259-2/2026/08}
\acmDOI{10.1145/3770855.3817672}

\begin{document}

\title{Ordering-based Causal Discovery via Generalized Score Matching}


\author{Vy Vo}
\affiliation{%
  \institution{Monash University}
  \city{Melbourne}
  \state{VIC}
  \country{Australia}
}
\email{Tran.Vo@monash.edu}

\author{Trung Le}
\affiliation{%
  \institution{Monash University}
  \city{Melbourne}
  \state{VIC}
  \country{Australia}
}

\author{He Zhao}
\affiliation{%
 \institution{The Commonwealth Scientific and Industrial Research Organisation}
 \city{Sydney}
  \state{NSW}
 \country{Australia}
 }

 \author{Edwin V. Bonilla}
\affiliation{%
 \institution{The Commonwealth Scientific and Industrial Research Organisation}
 \city{Sydney}
  \state{NSW}
 \country{Australia}
 }

\author{Dinh Phung}
\affiliation{%
  \institution{Monash University}
  \city{Melbourne}
  \state{VIC}
  \country{Australia}
}

\renewcommand{\shortauthors}{Vy Vo, Trung Le, He Zhao, Edwin V. Bonilla, \& Dinh Phung}

\begin{abstract}
Learning DAG structures from purely observational data remains a long-standing challenge across scientific domains. An emerging line of research leverages the score of the data distribution to initially identify a topological order of the underlying DAG via leaf node detection and subsequently performs edge pruning for graph recovery. This paper extends the score matching framework for causal discovery, which is originally designated for continuous data, and introduces a novel leaf discriminant criterion based on the discrete score function. Through simulated and real-world experiments, we demonstrate that our theory robustly enables accurate inference of true causal orders from observed discrete data and the identified ordering can significantly boost the accuracy of existing causal discovery baselines on most of the settings.
\end{abstract}


\maketitle

\input{inputs/1-Intro}
\input{inputs/2-Prelim}

\input{inputs/3-Review-SM}

\input{inputs/4-Method}

\input{inputs/5-Experiment}

\input{inputs/Conclusion}

\section{Acknowledgement}
Dinh Phung and Trung Le were supported by ARC DP230101176, DP250100262 and by the Air Force Office of Scientific Research under award number FA2386-23-1-4044. This does not imply endorsement by the funding agency of the research findings or conclusions. Any errors or misinterpretations in this paper are the
sole responsibility of the authors.

\bibliographystyle{ACM-Reference-Format}
\bibliography{bibs/sm,bibs/cdisc,bibs/stats,bibs/misc}

\appendix
\input{inputs/Proof}

\input{inputs/APDX-RWork}

\end{document}

%% file: math_commands.tex
\usepackage{amsmath,amsfonts,bm}

\def\eqref#1{equation~\ref{#1}}

\def\1{\bm{1}}

\def\rva{{\mathbf{a}}}
\def\rvb{{\mathbf{b}}}

\def\rvp{{\mathbf{p}}}

\def\rvr{{\mathbf{r}}}
\def\rvs{{\mathbf{s}}}

\def\rmE{{\mathbf{E}}}

\def\rmG{{\mathbf{G}}}

\def\rmM{{\mathbf{M}}}

\def\rmS{{\mathbf{S}}}

\def\rmV{{\mathbf{V}}}

\def\ermE{{\textnormal{E}}}

\DeclareMathAlphabet{\mathsfit}{\encodingdefault}{\sfdefault}{m}{sl}
\SetMathAlphabet{\mathsfit}{bold}{\encodingdefault}{\sfdefault}{bx}{n}

\def\gA{{\mathcal{A}}}

\def\gL{{\mathcal{L}}}
\def\gM{{\mathcal{M}}}
\def\gN{{\mathcal{N}}}
\def\gO{{\mathcal{O}}}
\def\gP{{\mathcal{P}}}

\def\gX{{\mathcal{X}}}

\def\sS{{\mathbb{S}}}

\newcommand{\E}{\mathbb{E}}

\newcommand{\Var}{\mathrm{Var}}

\newcommand{\Ch}{\mathrm{ch}_}
\newcommand{\Pa}{\mathrm{pa}_}
\newcommand{\Mb}{\mathrm{mb}_}
\newcommand{\De}{\mathrm{de}_}

\DeclareMathOperator*{\argmax}{arg\,max}
\DeclareMathOperator*{\argmin}{arg\,min}

%% file: inputs/1-Intro.tex
\section{Introduction}

Discovering the causal structure, often a \textit{directed acyclic graph} (DAG), underlying a system of variables has been an active pursuit across diverse scientific fields \cite{sachs2005causal,richens2020improving,wang2020causal}. This paper focuses on causal discovery from observational data, a central problem in causality that presents two key challenges.  First, identifiability remains a major issue: multiple causal models can generate the same observational data distribution. To this end, certain assumptions on the data generative process are required to ensure the causal model is identifiable from purely observed data \citep{peters2010identifying,peters2014causal,vo2024optimal}. 

Second, structure learning is computationally intractable in the general case, as searching over the combinatorial space of DAGs is known to be NP-hard \citep{chickering1996learning,chickering2004large}. An important fact one can possibly exploit is that any DAG imposes at least one topological order and the ordering exists if and only if it is a DAG. The prior knowledge of partial orderings is typically available in some real-world scenarios, such as genetics \citep{olson2006gene}, healthcare \citep{denton2007optimization} or meteorology \citep{bruffaerts2018comparative}.  Incorporating such prior information can significantly reduce the complexity of DAG search since acyclicity constraint is naturally enforced given a causal order \citep{bandifferentiable}. 

Ordering-based causal discovery is a line of research that addresses the case where orderings are not given \citep{teyssier2012ordering,buhlmann2014cam}. The algorithm entails two stages: (1) determining a topological order and (2) subsequent post-processing to remove spurious edges. Research in ordering-based causal discovery has recently taken off with the use of score matching \citep{rolland2022score,sanchez2022diffusion,montagna2023causal,montagna2023shortcuts,xuordering}, in which a valid causal order can be estimated by sequentially identifying the \textit{sink} or \textit{leaf} nodes (i.e., nodes with no outgoing edges) based on the \textit{score} of data distribution (i.e., Jacobian of the data log-likelihood). The approach has been shown to be practically effective as well as robust to noise misspecifications or assumptions violations such as faithfulness and measurement errors \citep{montagna2024assumption}.

Despite their successes, ordering-based causal discovery frameworks with score matching are currently limited to continuous data.  Extending the methods to discrete data remains a largely unexplored area. The core difficulty lies in the fact that the concept of a score function  is not well-defined for discrete random variables.  Our work is motivated by a fundamental question: \textbf{can the score matching paradigm be applied for recovering a causal order from discrete data?} Given the growing  literature on surrogate ``scores'' for discrete data \citep{hyvarinen2007some,lyu2012interpretation,meng2022concrete,sun2022score}, we investigate whether any of the proposed discrete score functions can effectively serve as a leaf node discriminant criterion. It turns out the answer is affirmative, and we further develop an identifiability result that guarantees the recovery of causal orders from observational discrete data.

\paragraph{\textbf{Contributions.}} Our contributions in this work can be summarized in the following: 
\begin{itemize}[leftmargin=*]
    \item We characterize the identifiability of a topological order underlying a discrete structural causal model and demonstrate how it can be estimated from the \textit{discrete score} of data distribution. To our best knowledge, this is the \underline{first score matching method} for learning the full causal order from discrete data.

    \item With no assumption about the additive structure, our theory gives rise to a condition that extends existing identifiability results for causal order in additive noise.

    \item We enrich the ordering-based literature with a generalization to discrete data, while providing a fresh view to learning structures from categorical variables, which is currently dominated by classical independence test-based and score-and-search approaches (see related works in Appendix \ref{sec:rwork}).

    \item Lastly, we validate our theory through synthetic and real-world experiments, showing empirical effectiveness and robustness in recovering true causal orders under noisy and relaxed conditions. 
\end{itemize}

%% file: inputs/2-Prelim.tex
\section{Preliminaries}

\paragraph{Notation.} We use upper case letters (e.g., $X$) for random variables and lower case letters (e.g., $x$) for values. We reserve bold capital letters (e.g., $\rmG$) for notations related to graphs and calligraphic letters (e.g., $\gX$) for spaces. Finally, we use $[d]$ to denote a set of integers $\{1, 2, \cdots, d\}$. 

\subsection{Structural Causal Model}
A directed graph $\rmG = (\rmV, \rmE)$ consists of a set of nodes $\rmV$ and an edge set $\rmE \subseteq \rmV^2$  of ordered pairs of
nodes with $(v, v) \notin \rmE$ for any $v \in \rmV$ (one without self-loops). For a pair of nodes $i,j$ with $(i,j) \in \ermE$, there is an arrow pointing from $i$ to $j$ and we write $i \rightarrow j$. Two nodes $i$ and $j$ are adjacent if either $(i,j) \in \rmE$ or $(j,i) \in \rmE$. If there is an arrow from $i$ to $j$, then $i$ said to be a parent of $j$ and $j$ is a child of $i$. Let $\Pa{i}$ and $\Ch{i}$ denote the set of variables respectively associated with parents and children of node $i$ in $\rmG$. A node $j$ is a descendant of $i$ in $\rmG$ if there exists a directed path from $i$ to $j$. 

A \textit{structural causal model} \citep[SCM,][]{pearl2009causality} characterizing the data generative process for a set of random variables $X = \{X_i\}_{i \in [d]}$ is defined over a tuple $\langle U, X, f \rangle$ that consists of a sets of assignments 
\begin{align}\label{eq:scm}
    X_i := f_i \big(X_{\Pa{i}}, U_i \big), \quad i \in [d], 
\end{align}
where  $U = \{U_i\}_{i\in[d]}$ are mutually independent exogenous variables with strictly positive density. Given a joint distribution over the exogenous variables $U$, the (deterministic) functions $\{f_i\}_{i \in [d]}$ define a joint distribution $P_X$ over the endogenous variables $X$. An SCM without feedback loops induces a DAG structure $\rmG$. An important property of DAGs is that there exists a non-unique \textit{topological order} $\pi = (\pi_1, \cdots, \pi_d)$ that represents directions of edges such that $i$ comes before node $j$ in the ordering for every directed edge $(i,j) \in \rmE$, which is often written as $\pi_i \prec \pi_j$, where $\pi_i, \pi_j$ denote the positions of nodes $i$ and $j$ in the ordering. In this work, we make standard causal discovery assumptions: (1) the distribution $P_X$ satisfies the Markov properties w.r.t $\rmG$ i.e., $X_i \indep X_j \mid X_{\Pa{i}}, \forall i\in[d], \forall j\in[d] \backslash (\De{i} \cup \Pa{i} \cup \{i\})$; and (2) there are no latent confounders among the observed variables. Let $p(x)$ be the joint probability density function over $X$. This model allows $p(x)$ to be factorized according to the structure $\rmG$ as $p(x) = \prod^{d}_{i=1} p(x_i \vert x_{\Pa{i}})$.

\subsection{Generalized Score Matching} \label{sec:gsm}

Score matching is a family of parameter learning methods alternative to the maximum likelihood principle. The objective entails matching two log probability density functions by their first-order derivatives, i.e., $\nabla \log p(x)$, using the Fisher divergence.  
We refer readers to the milestone literature on score matching in \citep{vincent2011connection,hyvarinen2005estimation,hyvarinen2007some,lyu2012interpretation,song2019generative}. In this section, we focus on the generalized score matching principle proposed in \citet{lyu2012interpretation}. We later show how the following generalization facilitates the identification of a valid topological ordering of $\rmG$ from discrete data.

Given two densities $p(x)$ and $q(x)$ and a linear operator (functional) $\gL$, the \textit{generalized Fisher divergence} is defined as 
\begin{align}
    D_{\gL}(p \Vert q) = \sum_{\gX} p(x) \Bigg\vert \frac{\gL p(x)}{p(x)} - \frac{\gL q(x)}{q(x)} \Bigg\vert^2,
\end{align}
where $\frac{\gL p(x)}{p(x)}$ is termed as \textit{generalized score function}. A valid linear operator $\gL$ should be complete, meaning that two densities $p(x) = q(x)$ (a.e) if $p(x)$ and $q(x)$ satisfies $\frac{\gL p(x)}{p(x)} = \frac{\gL q(x)}{q(x)}$ (a.e). It is easy to see that the gradient operator $\nabla$ is complete, under which $D_{\gL}$ reduces to the original Fisher divergence, since $\nabla \log p(x) = \frac{\nabla  p(x)}{p(x)}$.  

For discrete data, \citet{lyu2012interpretation} proposes to choose $\gL$ to be the marginalization operator $\gM$. Let $\gM_i p(x) := p(x_{-i})=\sum_{x_i} p(x)$ be the marginal density induced from $p(x)$, where $x_{-i}$ denote the vector formed by dropping $x_i$ from $x$. This gives rise to 
\begin{align}\label{eq:discrete-score}
    \frac{\gM_i p(x)}{p(x)} = \frac{p(x_{-i})}{p(x)} = \frac{1}{p(x_i \vert x_{-i})}.
\end{align}
The \textbf{discrete score function} is defined as $\gM p(x) := \left[\gM_i p(x)\right]^{d}_{i=1} $ where each $\gM_ip(x)$ is a reciprocal of the singleton conditional density $p(x_i \vert x_{-i})$. 

The operator $\gM$ is complete due to a well-known result in statistics \citep{brook1964distinction,lyu2012interpretation} that the joint density $p(x)$ is completely determined by the ensemble of the singleton conditionals $p(x_i \vert x_{-i}), \forall i\in[d]$.  Here the normalizing constant does not affect the computation as it gets cancelled out in the generalized score function. The generalized Fisher divergence can also be re-expressed into a form as an expectation of functions of the unnormalized model, which enables Monte Carlo sampling for estimation. It is worth noting that the above construction is also applicable to continuous data where the summation is replaced with integration.

%% file: inputs/3-Review-SM.tex
\section{Learning DAGs from the Score}

\begin{table*}[t!]
\centering
\caption{Evolution of causal discovery methods based on score matching.}
\label{tab:comparison}
\resizebox{0.85\linewidth}{!}{
\begin{tabular}{l | c c c c c c c}
\toprule[2pt]
Method & \texttt{LISTEN} \citep{ghoshal2018learning} & \texttt{SCORE} \citep{rolland2022score} & \texttt{DAS} \citep{montagna2023scalable}  & \texttt{NoGAM} \citep{montagna2023causal}  & \texttt{DiffAN} \citep{sanchez2022diffusion}  & \texttt{CaPS} \citep{xuordering}  & \texttt{AdaScore} \citep{montagna2024score} \\
\midrule[1pt]
Non-linear ANMs & \textcolor{red}{\xmark} & \cmark & \cmark & \cmark & \cmark & \cmark & \cmark\\
Linear ANMs  & \cmark & \textcolor{red}{\xmark} & \textcolor{red}{\xmark} & \textcolor{red}{\xmark} & \textcolor{red}{\xmark} & \cmark & \cmark \\
Non-Gaussian noises  & \cmark & \textcolor{red}{\xmark} & \textcolor{red}{\xmark} & \cmark  & \cmark & \textcolor{red}{\xmark} & \cmark \\
\midrule[1pt]
Non-additive mechanisms  & \textcolor{red}{\xmark} & \textcolor{red}{\xmark} & \textcolor{red}{\xmark} & \textcolor{red}{\xmark}  & \textcolor{red}{\xmark} & \textcolor{red}{\xmark} & \textcolor{red}{\xmark}\\
Discrete variables  & \textcolor{red}{\xmark} & \textcolor{red}{\xmark} & \textcolor{red}{\xmark} & \textcolor{red}{\xmark}  & \textcolor{red}{\xmark} & \textcolor{red}{\xmark} & \textcolor{red}{\xmark}\\
\bottomrule[2pt]
\end{tabular}
}
\end{table*}

In this section, we summarize the key theoretical developments driving advances in score-matching-based causal discovery literature. An important class of causal models for continuous data is additive noise model \citep[ANM,][]{peters2014causal,hoyer2008nonlinear} where (\ref{eq:scm}) takes the form 
$$X_i := f_i \big(X_{\Pa{i}}) + U_i, \quad \forall i \in [d].$$ The model is known to be identifiable from observed data if the functions $\{f_i\}_{i\in [d]}$ are twice continuously differentiable and non-linear in every component.

\texttt{SCORE} \citep{rolland2022score} is the pioneering work that sheds light on the connection between score function and causal discovery. Assuming the model is a non-linear ANM with Gaussian noise where the noise variables $U_i \sim \gN(0, \sigma^2_i)$, the authors provide a provable recovery of a valid causal order of the true DAG from the knowledge of the score function $\nabla \log p(x)$. 

By the Markovian factorization, the joint log density under this model can be written as 
\begin{align*}
    \log p(x) & =\sum_{i=1}^{d} \log p(x_i \vert x_{\Pa{i}}) \\
    & = -\frac{1}{2} \sum_{i=1}^{d} \left( \frac{x_i - f_i(x_{\Pa{i}})}{\sigma_i} \right)^2 - \frac{1}{2} \sum_{i=1}^{d} \log (2\pi\sigma^2_i). 
\end{align*}

Thus, the score function $\rvs(x)  := \nabla \log p(x)$ reads 
\begin{align} \label{eq:SCORE}
    \rvs_j(x) = - \frac{x_j - f_j(x_{\Pa{j}})}{\sigma_j^2} + \sum_{i \in \Ch{j}} \frac{\partial f_i}{\partial x_j}(\Pa{i}) \frac{x_i - f_i(x_{\Pa{i}})}{\sigma^2_i}. 
\end{align}

If $j$ is a leaf node, then the second summand vanishes due to having no children, which gives rise to 
$\partial_j \rvs_j(x) := \frac{\partial \rvs_j(x)}{\partial x_j} = -1 \slash \sigma^2_j$. The resulting leaf discriminant criterion is given as
\begin{align}\label{crit:var}
 X_j \text{ is a leaf } \Leftrightarrow \Var [\partial_j \rvs_j(x)] = 0.   
\end{align}

\texttt{LISTEN} \citep{ghoshal2018learning} can be derived as a linear case of \texttt{SCORE} where the leaf node can be read off as the entry corresponding to the minimum value of the diagonal of the precision matrix over the noise variances. Nonetheless, \texttt{LISTEN} or \texttt{SCORE} is not applicable to models with mixed (unknown) linear and non-linear relations. Specifically, 
\texttt{SCORE} cannot distinguish leaf nodes in linear ANMs, where all diagonal values of the score's Jacobian are constants.

Improving on \texttt{SCORE}, \citet{xuordering} propose \texttt{CaPS}, an alternative leaf discriminant criterion applicable to both linear and non-linear relations, where the outer variance is replaced with expectation. A sufficient condition for identifiability is that the noise variances are non-decreasing w.r.t any causal order $\pi$, adapted from Assumption 1 in \citet{xuordering} as follows:

\begin{definition}[Non-decreasing variance of noises] A topological order $\pi$ satisfies the non-decreasing variance property if for any two noise variables $U_i$ and $U_j$, one has $\sigma_i \le \sigma_j$ if $\pi_i < \pi_j$.
\end{definition}

This assumption also serves as an extension of the identifiability condition of linear ANMs based on (inverse) covariance matrix in \citet{ghoshal2018learning} (see Assumption 1 therein).  

In an arbitrary noise setting where the noise variables are i.i.d. with smooth density $p(u)$,  
the first term in (\ref{eq:SCORE}) takes a general form of $\partial_{u_j} \log p(u_j)$, which corresponds to the score of $X_j$ if $j$ is a leaf node, that is $\rvs_j (x) = \partial_{u_j} \log p(u_j), \forall j\in[d], \Ch{j} = \emptyset$. The authors of \texttt{NoGAM} \citep{montagna2023causal} capitalize on this property showing that the noise term of a leaf variable $X_l$ is equivalent to the residual defined as 
\begin{align*}
    U_l = R_l, \quad R_l := X_l - \mathbb{E}\left[X_l \vert X \backslash \{X_l\}\right],
\end{align*}
where $\mathbb{E}\left[X_l \vert X \backslash \{X_l\}\right]$ is the optimal least squares predictor of $X_l$ by the all remaining nodes. Since the score $\rvs_l(x)$ is a function of $R_l$, one has the following leaf discriminant criterion
\begin{align}\label{crit: residual}
 X_j \text{ is a leaf } \Leftrightarrow \mathbb{E} \left[ \left( \mathbb{E} \left[ \rvs_j(x) \vert R_j \right] - \rvs_j(x) \right)^2 \right] = 0.   
\end{align}
 
Built on the above result, \texttt{AdaScore} \citep{montagna2024score} extends \texttt{NoGAM} to linear cases and further accommodates the presence of latent confounders. 

Evidently, two necessary conditions underlie the above frameworks. One is that the true SCM takes the additive form with independent noise terms; otherwise the noise effect of each variable cannot be separated from that of its children and co-parents. Another condition is  no local density is degenerate so that not only is the score function well-defined but also no non-leaf node satisfies criteria (\ref{crit:var}) or (\ref{crit: residual}) trivially. This is formalized in Condition \ref{ass:nonzero}.

\begin{condition}[Non-degeneracy]\label{ass:nonzero}
    Let $x \in \gX$ be a random vector defined via an SCM (\ref{eq:scm}). For any node $i \in [d]$, the conditional densities $p(x_i \vert x_{\Pa{i}})$ are non-zero $\forall x \in \gX$. 
\end{condition}

Table \ref{tab:comparison} presents an overview of the existing approaches in this line of research and reveals an unexplored question of handling discrete data in non-additive noise models. \textbf{This motivates our main contribution: establishing the identifiability of causal order with a discrete score function.}

%% file: inputs/4-Method.tex
\section{Ordering-based Causal Discovery via Discrete Score Matching}

From this point we will deal with categorical random variables $X$ of finite domain where each variable $X_i$ has $n_i$ states $(n_i \ge 2)$ and its domain is $[n_i]$. Let $\gX := \prod_{i=1}^{d} [n_i]$ denote the domain of $X$. and $p(x)$ again be the joint probability density function. The key leverage is the singleton conditionals  $p(x_i \vert x_{-i})$, referred to as the \textbf{reciprocal discrete score functions}.

Our task is to identify a criterion to discriminate leaf nodes of a causal graph from i.i.d observational categorical data. The motivation of our theory begins with the \textit{non-decreasing variance} condition, which can be regarded as a type of prior knowledge about the uncertainty inherent in the system. We investigate whether such knowledge can facilitate the identification of the leaf variables in a discrete SCM.

Translation of the non-decreasing variance condition to discrete variables is however not straightforward, as they lack inherent quantitative values that can directly reflect the system's uncertainty. As revealed shortly, there fortunately exists a broad class of randomness measures of discrete probability distributions that only deals with the probabilities rather than the values on the associated sample space. Building on this construction, we develop a generalized framework for characterizing the system’s randomness. This framework plays a crucial role in identifying the leaf variable with the reciprocal discrete score function.

Let $\gP$ denote a class of all discrete probability vectors. We assume that all the probability vectors we deal with have been ordered in non-increasing order, and the vectors have an equal length of $n = \max(n_1, \cdots, n_d)$ by properly padding the
shorter one with the appropriate number of 0’s at the end.

\begin{definition}[Majorization]\label{def:majorization}
    Given two probability distributions $\rva = (a_1, \cdots, a_n)$ and $\rvb = (b_1, \cdots, b_n)$ with $a_1 \ge, \cdots, \ge a_n \ge 0$ and $b_1 \ge, \cdots, \ge  b_n \ge 0$, we say that $\rva$ majorizes $\rvb$, written as $\rva  \succeq \rvb$, if and only if 
        $\sum^{k}_{i=1} a_i \ge \sum^{k}_{i=1} b_i, \quad \text{for all } k = 1, \cdots, n.$
\end{definition}

\citet{hickey1982note,hickey1983majorisation} formalizes the randomness or spreadness of a discrete probability distribution via majorization theory \citep{marshall1979inequalities}.  For two discrete distributions $\rva$ and $\rvb$, we say $\rva$ is more spread or more uniform/random than $\rvb$ if $\rva \preceq \rvb$. A function $\phi: \mathbb{R}^{n} \mapsto \mathbb{R}$ is a \textit{Schur-concave} function if $\phi(\rva) \le \phi(\rvb)$ for all vectors $\rva, \rvb \in \mathbb{R}^n$ such that $\rva \succeq \rvb$. An interesting fact is the uniform vector i.e., $(\frac{1}{n}, \cdots, \frac{1}{n})$ is majorized by all probability vectors (thus being most random) and a degenerate vector e.g., $(1, 0, \cdots 0)$ and its permutations majorize all other vectors (thus being least random). These properties motivate the construction of a randomness measure as follows:

\begin{definition}[Measure of randomness \citep{hickey1982note}] \label{def:rand-measure} 
    A real-valued continuous function $\phi$, taking finite values in $\gP$ is a measure of randomness if it is symmetric and concave, and the concavity being strict on the sub-class of distributions with a finite number of positive probabilities. 
\end{definition}

Let $g: \mathbb{R} \mapsto \mathbb{R}$ be any continuous, strictly concave functions with $g(0)=0$. A popular class of such measures has the form:
\begin{align}\label{def:rand-form}
    \phi(\rvp) = \sum^{n}_{k=1} g \left(p_k \right).
\end{align}
The Shannon entropy function is one of the best-known measures of the above form, where $g(p_k) = - p_k\log p_k$. 

Finally, we impose a mild regularity condition, typically satisfied by well-behaved SCMs, requiring that every non-leaf node receive strictly additional information from its children and co-parents beyond its parents. This rules out degenerate conditional independencies and ensures that only leaf nodes have no extra information available beyond their parents.

\begin{condition}[Strict Markov blanket gain]\label{ass:strict}
For every non-leaf node $i$, let $\Mb{i}$ denote the Markov blanket and $b_i:=\Mb{i}\setminus \Pa{i}$. Assume there exists a set
$\gA_i\subseteq \mathrm{supp}(X_{\Pa{i}})$ with
$\mathbb \Pr(X_{\Pa{i}}\in \gA_i)>0$ such that for every
$x_{\Pa{i}} \in \gA_i$, there exist $x_{b_i}\neq x'_{b_i}$ with 
$$p(x_{b_i}\mid x_{\Pa{i}}), \qquad p(x'_{b_i} \mid x_{\Pa{i}})>0$$ 
and 
$$p(X_i \mid x_{\Pa{i}}, x_{b_i})\neq p(X_i \mid x_{\Pa{i}}, x'_{b_i}).$$
\end{condition}

\subsection{Identification of Causal Order}

With a slight abuse of notation, let $\phi(X)$ denote the randomness, under $\phi$, in the probability vector of the distribution of $X$. Let $X$ and $Y$ be jointly distributed discrete random variables. The conditional information is defined as $\phi(X \vert Y) = \mathbb{E}_{Y}\left[ \phi(X \vert y) \right]$, accordingly in the probability vector $p(X \vert y)$ for a given value $y$. 

\begin{condition}[Non-decreasing randomness]\label{ass:non-dec} 
For a measure of randomness $\phi$ defined in (\ref{def:rand-measure}), $\phi$ is said to satisfy the non-decreasing randomness condition w.r.t the true graph $\rmG$ if 
$$\phi(X_i \vert X_{\Pa{i}}) \le \min_{j\in \Ch{i}} \phi(X_j \vert X_{\Pa{j}}), \forall i \in [d], \Ch{i} \ne \emptyset.$$
\end{condition}

Condition \ref{ass:non-dec} characterizes along a causal order the relative uncertainty among local generative mechanisms. Upon closer examination, the condition essentially reflects the residual uncertainty introduced by environmental (exogenous) factors in the generation of a variable. From this view, one may find the presence of the property in realistic physical systems. Consider a thermodynamic example of an ice cube melting in a warm room. The parent state -- the ice -- has a highly ordered molecular structure. As the room temperature rises and the ice melts, the molecules transition to a more disordered liquid state. The child state -- the water -- thus has significantly higher uncertainty in its microstate than the original structured configuration of the ice. Or take an encryption example: one start with a deterministic plain message (the parent), and the child is an encrypted ciphertext (e.g., ``GJXPK'') generated by XORing the message with a random key. More generally, Condition \ref{ass:non-dec} tends to occur in cases where the child is a noisy effect of a clean parent signal, and its children continue to accumulate and amplify noises (thus randomness) through their generation.


We are now ready to state our identifiability results.

\begin{theorem}\label{theorem:main} 
    Let $x \in \gX$ be a discrete random vector defined via an SCM (\ref{eq:scm}), and let $\rvr_i(x_{-i}) := p(X_i \vert x_{-i})$ be the reciprocal discrete score function for every node $i \in [d]$. If there exists a randomness measure $\phi$ satisfying the non-decreasing randomness property w.r.t the true graph $\rmG$, then
    $X_j \text{ is a leaf node } \Leftrightarrow j \in \argmax_{i\in [d]} \mathbb{E}_{X_{-i}} \Big[ \phi\big( \rvr_i(x_{-i}) \big) \Big]$.
\end{theorem}
We say that the leaf variable $X_l$ is $\phi$-identifiable if Theorem \ref{theorem:main} holds for a certain measure $\phi$. The proof is delayed to Appendix \ref{apdx:proof-thm}. Note that our result implicitly assumes Condition \ref{ass:nonzero} to ensure that the singleton conditional densities are defined. 


Given any valid  measure $\phi$, the non-decreasing randomness condition w.r.t $\phi$ ensures that the leaf variables are $\phi$-identifiable in any subgraph of $\rmG$. Theorem \ref{theorem:main} suggests that this single condition is sufficient to recover a correct topological order of $\rmG$ through a sequential leaf node detection, which goes as follows: the scores $\rvr(\cdot)$ are initially estimated from observed discrete samples; then it selects as a leaf node the $\argmax$ of the criterion and the column corresponding to the selected leaf variable is then removed from the data matrix. The process is repeated on the new data until the entire ordering is determined. With accurate estimation of the scores, a valid causal order can ultimately be identified. Our causal order search procedure is detailed in in Algorithm \ref{algo:main}.

\paragraph{\textbf{Estimation of Discrete Score Function.}}
We employ continuous-time discrete diffusion models proposed in \citet{sun2022score} to estimate the singleton conditionals. The framework generalizes the ratio matching objective for binary variables from \citet{hyvarinen2007some}. The objective elegantly circumvents the calculation of the data score function and its minimizer is shown to be consistent. As for the parameterization of the score function, \citet{sun2022score} introduces a Transformer architecture that only requires $O(1)$ forward evaluations, which is adopted in our implementation. The model is designed in an amortized fashion where an entire ensemble of singletons is returned per input. In our implementation, models are trained on $4$ RTX $6000$ GPU cores with Adam optimizer \citep{kingma2014adam} at fixed $300$ epochs, $3000$ time steps and learning rates of $0.001$. The size of hidden units is set as $2d$ where $d$ is the number of variables in the data. For details on architecture design and the categorical ratio matching objective , we refer readers to Sections 4.1 and 5.3 in \citet{sun2022score}.

\input{inputs/algo1}

\subsection{Connection to Additive Noise Models} In relation to the previous literature, it is natural to ask whether the (expected conditional) variance function is applicable. 

Let us consider the function $\Var(\rvp) = \sum^{n}_{k=1} p_k (\log p_k - \mu)^2$ with $\mu=\sum^{n}_{k=1} p_k \log p_k$. The variance function is convex in the variables and in the presence of symmetry, convexity implies Schur-convexity. Hence, $\Var(\rvp)$ is Schur-convex, thus its negative, defined as $\phi_{\mathrm{Var}}(\rvp) := - \Var(\rvp)$ is Schur-concave and qualifies as a randomness measure. The variance function can thus be used for causal order search. This result is formalized in Corollary \ref{corl:main_var}, which simply follows from the result in Theorem \ref{theorem:main}.

\begin{corollary}\label{corl:main_var} 
 Let $x \in \gX$ be a discrete random vector defined via an SCM (\ref{eq:scm}), and let $\rvr_i(x_{-i}) := p(X_i \vert x_{-i})$ be the reciprocal discrete score function for every node $i \in [d]$. If $\phi_{\mathrm{Var}}$ satisfying the non-decreasing randomness property w.r.t the true graph $\rmG$, then
    $X_j \text{ is a leaf node } \Leftrightarrow j = \argmin_{i\in [d]} \mathbb{E}_{X_{-i}} \Big[ \Var\big( \rvr_i(x_{-i}) \big) \Big]$.
\end{corollary}

The proof is direct from Schur-convexity of the $\Var(\rvp)$. Furthermore, one may notice that in additive noise models, the uncertainty of the system is entirely captured in the noise variables. In this case, non-decreasing randomness of the local densities is reduced to non-decreasing randomness of the corresponding noise variables. If $\phi$ is the variance function, our Condition \ref{ass:non-dec} can be viewed as a generalization of the homoscedastic case and non-decreasing variance of noises introduced in the earlier literature.

A persistent challenge in causal discovery is the reliance on assumptions that are difficult to verify from data. In particular, popular parametric ANMs require checking for the additive structure and the distributions of the exogenous variables, which are rarely known nor directly testable in practice. Our theory addresses this testability gap in two complementary ways.  

First, with Condition \ref{ass:non-dec} subsuming ANMs, as soon as an ANM specification is available from domain knowledge, one can explicitly verify the condition  by measuring the randomness of the noise variables. Second, our theory also gives rise to another verification insight: it pinpoints special regimes where Condition \ref{ass:non-dec} can be analytically tested based on a structural priori about the in‑degree and cardinalities of a variable, its parents, and co‑parents. Proposition \ref{prop:main} makes this precise for a simple case of a symmetric Dirichlet prior w.r.t the Shannon entropy function. 

Formally, for each variable $X_i$ in a fixed DAG $\rmG$, let $\gX_{\Pa{i}} := \prod_{k \in \Pa{i}} [n_k]$ denote the set of configurations of the parent set of $X_i$. Let $K_i = |\gX_{\Pa{i}}|$ be the number of the configurations (i.e., CPT rows). Consider a discrete SCM in which each conditional distribution receives a symmetric Dirichlet prior with a common equivalent sample size $\alpha_0 > 0$, allocated uniformly across states. That is, for all $i \in [d]$ and $x_{\Pa{i}} \in \gX_{\Pa{i}}$:
\begin{align}\label{eq:scm-dir}
  p(X_i \mid x_{\Pa{i}}) \sim \mathrm{Dirichlet} \left( \frac{\alpha_0}{n_i K_i} \1_{n_i}\right).  
\end{align}

\begin{proposition}\label{prop:main}
Let $x \in \gX$ be a discrete random vector defined via a symmetric Dirichlet SCM (\ref{eq:scm-dir}). For sufficiently large $\alpha_0$, Condition \ref{ass:non-dec} holds w.r.t the Shannon entropy function $H(\cdot)$ if $\forall i \in [d]$ and $\forall j \in \Ch{i}$,
\begin{align}
\log \left( n_j \slash n_i \right) 
+ \alpha^{-1}_{0} \left[ K_i c(n_i) - K_j c(n_j) \right] \ge 0,
\end{align}
where $ c(n) := \frac{3}{2}n + \frac{1}{2} $.
\end{proposition}

Given knowledge of the graph’s in-degree and variable cardinalities, one can thus obtain a concrete, checkable criterion for our identifiability condition  without appealing to additive structure, functional form or noise law. We note that tractable approximations are available for other measures (e.g., Rényi entropies or variance) under more general Dirichlet models, and our proof template in Appendix \ref{apdx:proof-prop} can be well adapted to these other settings.

%% file: inputs/algo1.tex
\begin{algorithm}[t!] 
    \caption{Causal Order Search with Discrete \texttt{SCORE}}
\label{algo:main}
\begin{algorithmic}
\State \textbf{Input:} Data matrix $X \in [n]^{N \times d}$ and base measure $\phi$. 

\State \textbf{Output:} Topological ordering $\pi$.

\State Initialize $\pi \gets [\quad]$, nodes  $\gets \{1, \cdots, d\}$

\For{$i = 1, \cdots, d$} 

    \State Estimate the set of conditionals  $\big\{p(X_j \vert x_{-j})\big\}_{j \in \text{nodes}}$ 

    \State by training a continuous-time diffusion model;
    
    \State Estimate $V_j = \mathbb{E} \Big[ \phi \big( p(X_j \vert x_{-j})\big)\Big], \forall j \in \text{ nodes} $;
    
    \State Find leaf $l \gets \text{ nodes }[\argmax_j V_j]$;
    
    \State Update $\pi \gets [l, \pi]$, \quad $\text{nodes } \gets \text{ nodes } - \{l\}$;
    
    \State Remove $l$-the column of $X$.
\EndFor
\end{algorithmic}
\end{algorithm}

%% file: inputs/5-Experiment.tex
\section{Numerical Experiments}

\subsection{Experimental Design}

\paragraph{\textbf{Datasets.}}

We empirically validate the theory through numerical experiments on synthetic and real-world datasets. We first simulate random DAGS of \textit{Erdos-Rényi} (ER) and \textit{Scale-Free} (SF) structures with number of nodes $d$ up to $60$ and expected degrees at $2d$ and $4d$. We populate the conditional probability tables with normalized uniformly random weights. This strategy aims to create a system of approximately constant randomness (from standard uniform exogenous noises), enabling the verification of our identifiability results. The cardinality of variables runs from $3$ to $6$. 

We additionally experiment with six real-world discrete Bayesian networks in the \texttt{bnlearn} repository\footnote{\url{bnlearn.com} \& \url{pgmpy.org}}: \textit{Earthquake} (EQ, $d=5$), \textit{Sachs} (SA, $d = 11$), \textit{Child} (CH, $d = 20$), \textit{Insurance} (IS, $d = 27$), \textit{Mildew} (MD, $d = 35$) and \textit{Alarm} (AL, $d = 37$). In each setting, we generate $10$ random datasets of $N = 10,000$ observations are sampled from the given models. 
We investigate the variance function and Shannon entropy as randomness measures. We select the measures yielding the least risk of violations on respective datasets based on the proposed diagnosis. We find that the former works best in all simulations, while the latter performs optimally on the remaining networks. We also provide implementation for the other options based on the Rényi entropy and $\mathrm{KL}$ divergence in our codes\footnote{available at \href{https://github.com/isVy08/discrete-SCORE}{github.com/isVy08/discrete-SCORE}}.  

\paragraph{\textbf{Baselines.}}
We investigate representative algorithms of $3$ families of causal discovery approaches for discrete data: (i) constraint-based methods with the \texttt{PC} algorithm \citep{spirtes1991algorithm}, (ii) score-and-search methods with the classic \texttt{GES} \citep{chickering2002optimal} and \texttt{ORDCD} \citep{ni2022ordinal} -- a recent algorithm for ordinal data, and (iii) generalized additive models \citep[\texttt{GAM},][]{wood2017generalized}. We report the performance of the \texttt{PC} algorithm with $\mathsf{G}$-tests \citep{quine1985efficiencies}. For \texttt{GES}, we test both BIC and BDeu scores while for \texttt{ORDCD}, we adopt the authors' default setup that uses BIC score for greedy search. Lastly, we explore additive models for variable selection by fitting a multinomial logistic regression model with factor covariates. It eliminates redundant edges from a fully connected graph based on a cut-off value of $0.001$.

\paragraph{\textbf{Metrics.}} The quality of the estimated order can be assessed by how well the provided knowledge from the ordering can improve causal discovery baselines. For DAG evaluation, we compute the commonly used metrics: structural Hamming distance (SHD), structural Intervention distance (SID) and average F1 score of skeleton F1 and directionality F1, measuring adjacency and orientation accuracies respectively. SHD counts the number of single-edge mistakes: extra, missing, or reverse, with each error unit counted once per unique node pair. SID provides a complementary perspective by evaluating whether the learned graph supports reliable causal inference under interventions. It measures the number of interventional distributions that are falsely inferred by the estimated causal graph w.r.t the true one.  While SHD favours sparse graphs, SID penalizes overly sparse ones, resulting in a more balanced and comprehensive performance evaluation. We further report $\mathrm{D}_{top}$, a topological divergence metric proposed by \citet{rolland2022score} for quantifying the number of edges that \underline{cannot} be recovered due to the errors in the topological order. For an ordering $\pi$ and a target adjacency matrix $A$, the metric is defined as $\mathrm{D}_{top}(\pi, A) = \sum^{d}_{j=1} \sum_{i: \pi_j > \pi_i} A_{ji}.$

$\mathrm{D}_{top}(\pi, A)$ returns zero if $\pi$ is a correct order. It provides a lower bound on the SHD of the final algorithm (irrespective of the pruning method). For comparative purposes, we report the normalized values of SHD and SID over the number of edges. Higher F1 ($\uparrow$) and lower $\mathrm{D}_{top}$, SHD, SID ($\downarrow$) and are desirable.

\paragraph{\textbf{Post-processing for DAG Estimation.}}Once a causal order is found, a post-processing step is applied for recovering the DAG. Methods for continuous data often rely on regression to identify parent variables, which requires knowledge of the appropriate model forms. However, our approach does not assume an additive structure, making it more flexible yet more challenging in terms of selecting the suitable pruning method. We here introduce two approaches to applying the prior knowledge from an inferred causal order on top of an existing structural learning algorithm.

The first strategy is \textbf{(1) edge pruning}: if a node $j$ appears after node $i$ in the true ordering, i.e., $\pi_j > \pi_i$, the edge $j \rightarrow i$ cannot exist; as a result, the inferred ordering introduces the forbidden links from a node to its preceding nodes, which can be used to impose constraints on DAG search algorithms. This strategy is most effective when the base method has high recall but low precision, which is observed in \texttt{PC} and \texttt{GAM} algorithms. In this case, the removal of forbidden edges helps decrease false positives, leading to a boost in F1 scores to these methods.

Meanwhile, score-and-search methods such as \texttt{GES} tends to give higher precision at the cost of recall. Given a correct ordering, pruning edges would increase precision with no impact on recall, thereby increasing F1 accuracy. However, if the ordering is imperfect, edge deletion would often worsen F1 by increasing precision while adversely affecting recall. Therefore, a reverse strategy -- \textbf{(2) edge insertion}, is found to be more effective when we deal with such methods, when it improves the detection of true positives by predicting more causal edges rather than removing them. While the causal direction cannot be determined from a topological order, we know that the parent nodes always precede their children in the ordering. Recall that \texttt{GES} searches for the true DAG based on a loss or scoring function -- often BDeu score \citep{heckerman1995learning} in discrete cases -- assuming the true DAG corresponds to the minimizer of the objective. Consider a node in the ordering under analysis, we propose to iteratively add a causal edge from its predecessor and evaluate the change in the score of the graph. The edge is accepted as long as such an insertion reduces the loss and does not create cycles, or in other words, it maintains acyclicity. Empirically, this strategy is found to help the algorithm escape local minima and search for more accurate graphs.

\input{inputs/fig-main}

\subsection{Results \& Discussion} 

It is worth emphasizing that our simulations are deliberately designed to be noisy, without enforcing any specific randomness measure or exact monotonicity in the true causal order. Likewise, the real-world settings provide no guarantee that Condition \ref{ass:non-dec} holds, and even there is a substantial chance that the condition is violated. Thus, the target of the evaluation is robustness: \textbf{given an imperfectly estimated order, to what extent is it helpful to guide causal discovery baselines toward the correct solution?} 

Figure \ref{fig:main-ER} to \ref{fig:main-REAL} present the key results in recovering the topological order and the true causal graph across different graph structures and sizes (number of nodes $d$)\footnote{We encounter a memory explosion issue in the data generation process for SF graphs as they tend to be concentrated on high-degree nodes. The experiments on SF graphs are therefore conducted up to $20$ nodes due to our memory constraints. }. \texttt{SCORE+[X]} refers to the application of the estimated causal orders on a baseline \texttt{[X]}\footnote{We only evaluate \texttt{SCORE} on high-performing base algorithms, which are coloured.}. In the main text, the reported settings have fixed sample size of $10,000$, and models are evaluated over $10$ random initializations.

Our method is effectively placed under a stress test, yet the empirical results are surprisingly strong under these loose conditions. Across all simulated settings and most real-world networks, our algorithm, the \texttt{SCORE+[X]} version, consistently improves upon existing baselines in terms of F1 accuracy and SID, with the latter directly reflecting higher reliability for causal inference. In many cases, SHD is slightly higher, which however does not necessarily equate to poor performance. Concretely, SHD can be misleading especially if the true causal graphs are sparse: there are far more opportunities to add an incorrect edge than to miss a true one, so even a few extra edges can inflate SHD substantially. As a result, SHD implicitly encourages algorithms to under-predict edges, since predicting too few is ``safer'' than predicting too many. Thus, a practically effective method should aim to maintain SHD or with minimal increase only while achieving high F1 and strong SID performance. This is  the behaviour we observe with our method. The insight is that while the condition may not hold globally, as long as it holds for any local generative system (i.e., a local family of parent-child variables), the algorithm can recover the correct sub-order. This explains why even partially correct order estimates remain sufficiently informative to improve downstream causal discovery.

\paragraph{Failure Modes.} Since the experiments operate in noisy regimes, adverse scenarios do occur, notably on \textit{Sachs} (SA) and \textit{Insurance} (IS) datasets. Two main factors contribute to cases where performance degrades: (1) estimation accuracy and (2) assumption validity -- the extent to which the monotonic pattern holds in the data w.r.t the selected randomness measure. When the sample size is sufficient large, the overall performance is substantially affected by the latter. An implication from our theory is that the inferred order would be more erroneous (more nodes are misplaced) when upstream nodes have higher uncertainty -- for example, when exogenous variation dominates over parental effects. The consequences are more severe to algorithms like \texttt{PC} where edge deletion is applied, ignoring many true edges. Meanwhile, for methods like \texttt{GES}, the impact is less so because the edge addition strategy also incorporates the judgment from the base algorithm, helping to mitigate any incorrect decisions.

\input{inputs/APDX-order}

%% file: inputs/fig-main.tex
\begin{figure*}[h!]
\centering
\begin{subfigure}[b]{0.90\textwidth}
   \includegraphics[width=\linewidth]{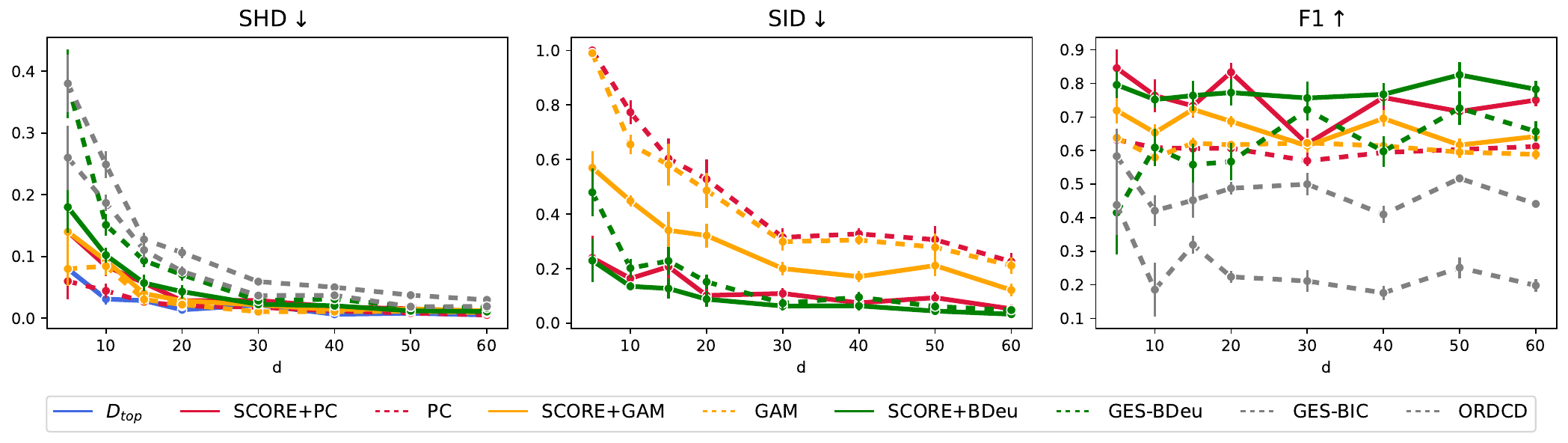}
\end{subfigure}
\begin{subfigure}[b]{0.90\textwidth}
   \includegraphics[width=\linewidth]{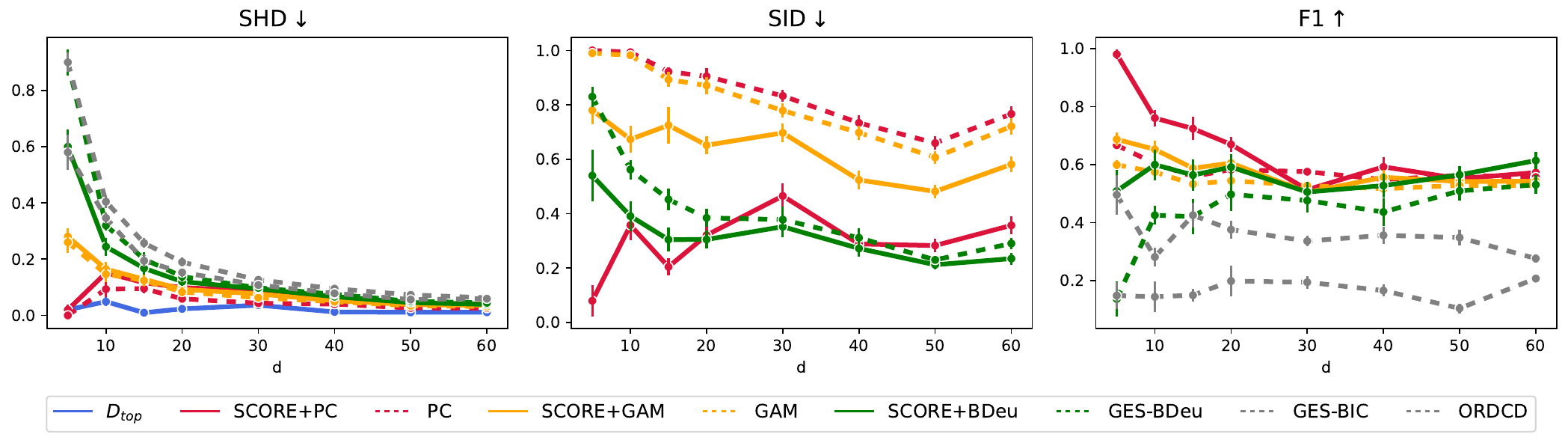}
\end{subfigure}
\vspace{-0.5em}
\caption{Experiments with ER graphs of  $2d$ \textbf{(top)} and $4d$ degrees \textbf{(bottom)}. $\texttt{SCORE+[X]}$ corresponds to our methods.}
\label{fig:main-ER}
\end{figure*}

\begin{figure*}[h!]
\centering
\begin{subfigure}[b]{0.90\textwidth}
   \includegraphics[width=\linewidth]{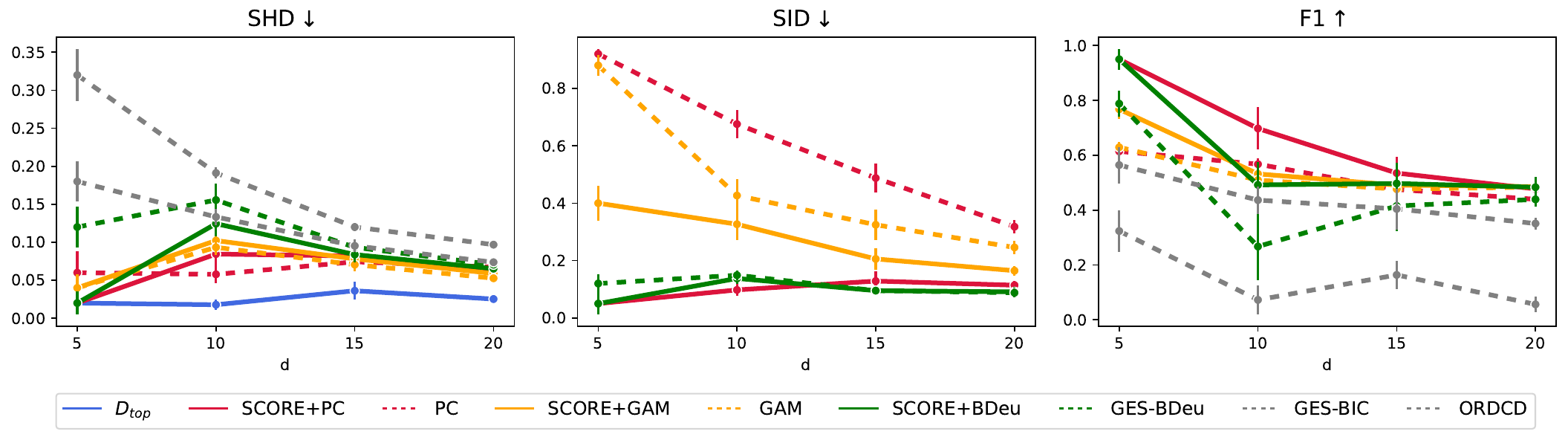}
\end{subfigure}
\begin{subfigure}[b]{0.90\textwidth}
   \includegraphics[width=\linewidth]{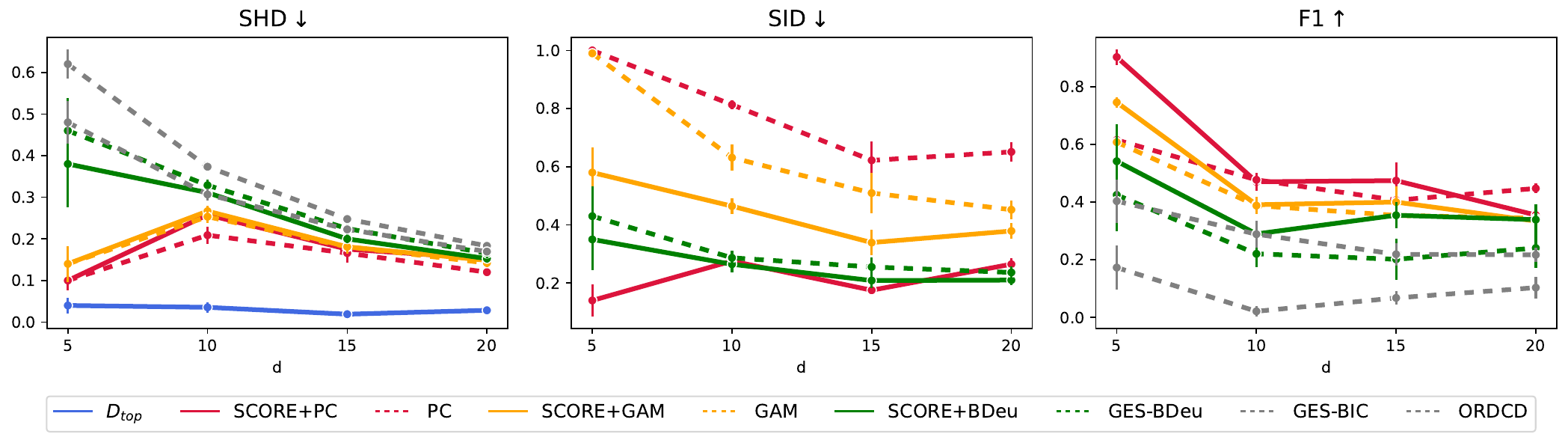}
\end{subfigure}
\vspace{-0.5em}
\caption{Experiments with SF graphs of $2d$ \textbf{(top)} and $4d$ degrees \textbf{(bottom)}. $\texttt{SCORE+[X]}$ corresponds to our methods.}
\label{fig:main-SF}
\end{figure*}

\begin{figure*}[!hbt]
    \centering
    \includegraphics[width=0.90\linewidth]{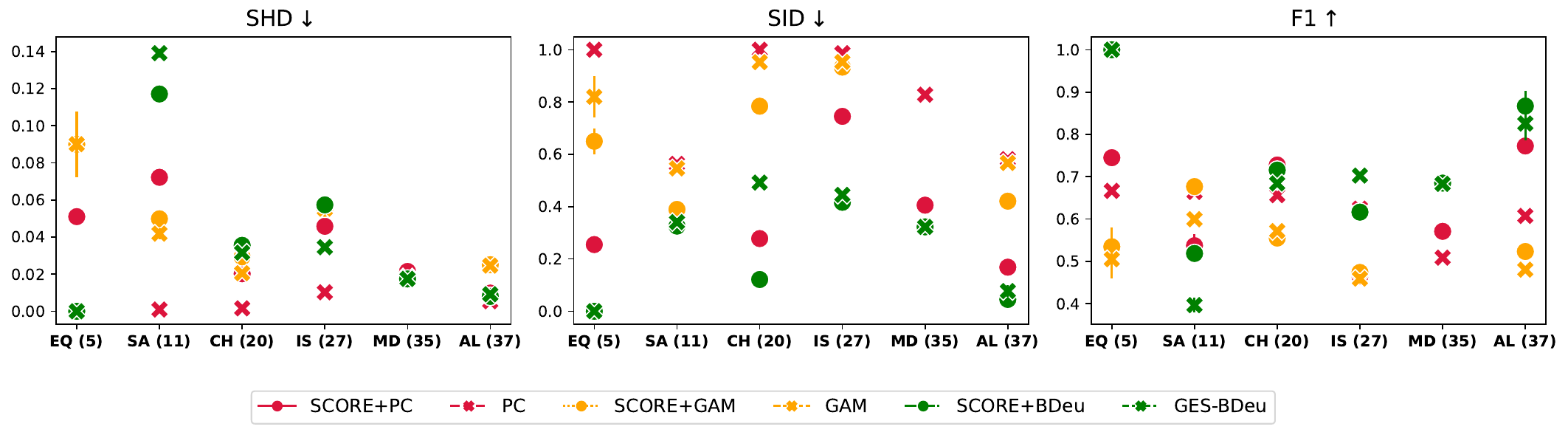}
    \vspace{-1em}
    \caption{Experiments with real-world networks on $\mathbf{10,000}$ samples. $\texttt{SCORE+[X]}$ corresponds to our methods}
    \label{fig:main-REAL}
\end{figure*}

\begin{figure}[!h]
    \centering
    \includegraphics[width=0.75\linewidth]{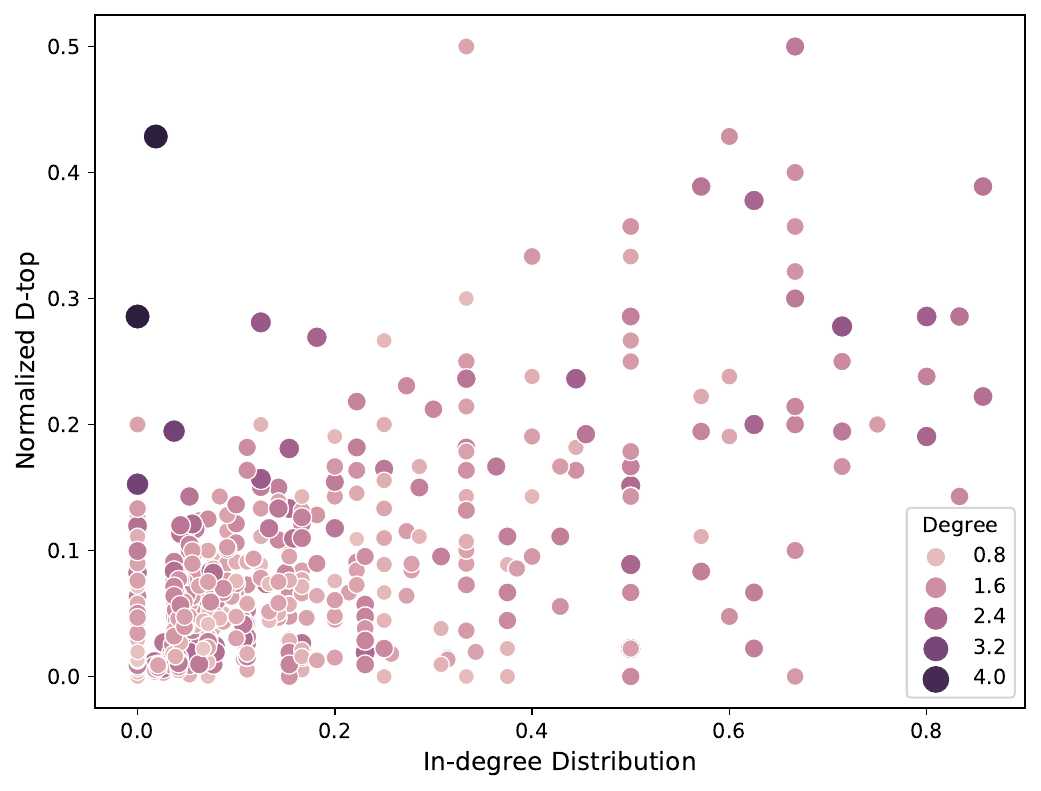}
    \vspace{-1em}
    \caption{Correlation between $\mathsf{D}_{top}$ and the in-degree distribution given $\mathrm{deg}^{-}_{\max} = 3$ across various graph degrees.}
    \label{fig:corr}
\end{figure}

\begin{figure}[!h]
    \centering
    \includegraphics[width=0.75\linewidth]{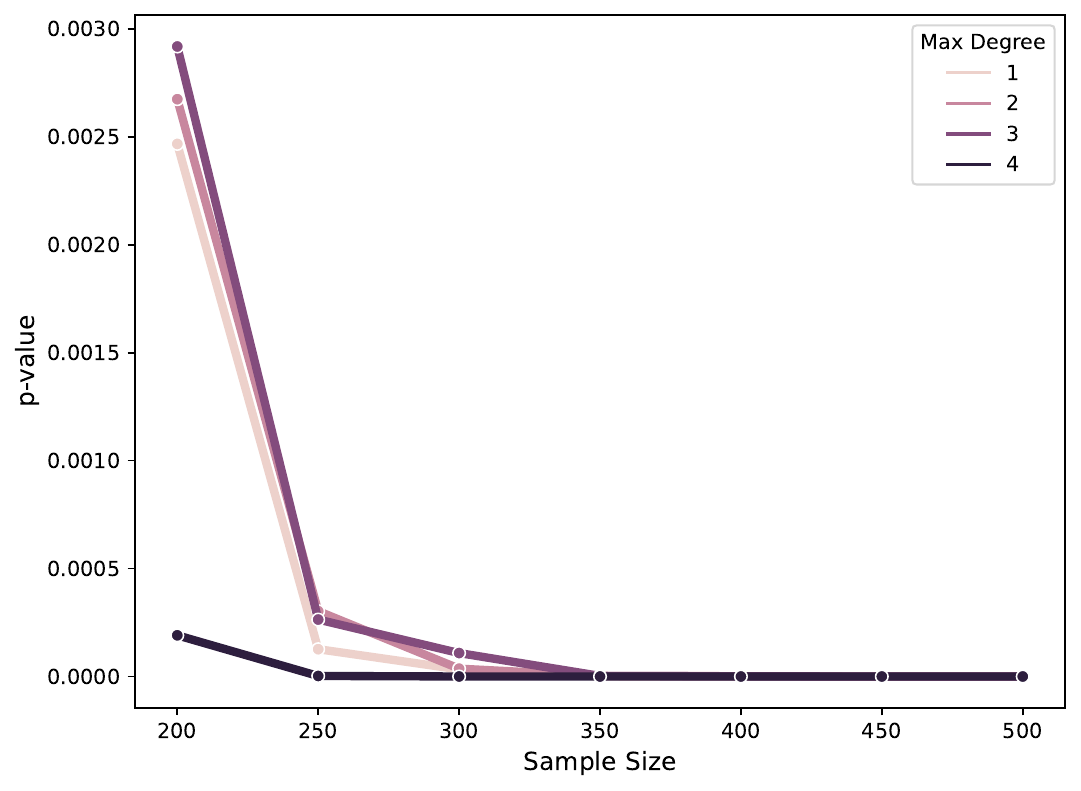}
    \vspace{-1em}
    \caption{$p$-values of KCI tests across sample sizes and hypotheses of $\mathrm{deg}^{-}_{\max}$.}
    \label{fig:pvalue}
\end{figure}


%% file: inputs/APDX-order.tex
\section{Practical Diagnosis for Order Validity}\label{apdx:markov} 

While Proposition \ref{prop:main} provides an elegant, structure-driven test of Condition \ref{ass:non-dec}, it still rests on (though widely used) parametric assumptions. This raises a natural question: how should practitioners proceed in the absence of valid parametric models? Our second contribution targets this challenge: we propose a practical diagnostic test for whether a topological order inferred from an algorithm is potentially invalid. Since most, if not all, conditions are untestable from data in full generality, heuristic tests are not uncommon. \citet{hoyer2008nonlinear} and \citet{peters2014causal} introduce a procedure to check for independent noise terms in bivariate additive noise models with regression and independence testing, in which violations indicate either model misspecifications or inconclusive causal direction. Our proposed approach also makes use of independence tests, which is arguably inevitable without additional assumptions.

We make use of mild knowledge about the sparsity of the underlying DAG $\rmG$: maximum in-degree, to diagnose the validity of a topological order. By Markov property, a node is independent of its non-descendants conditioned on the parent nodes. Given an estimated ordering $\pi$, we begin with Algorithm \ref{algo:markov} where we compute the in-degree distribution induced by $\pi$. Concretely, for each variable $X_j$, we consider the set $\sS$ of its predecessors and evaluate whether $X_j \indep X_i \mid \mathbb{S} \backslash \{X_i\}$ for all  $X_i \in \mathbb{S}$.  By definition, a valid topological order is one such that $i$ precedes $j$ if there is an edge $i \rightarrow j$ in $\rmG$. Since $\rmG$ satisfies Markov property, every $X_j$ is independent of the non-descendants conditioned on its parent variables. Therefore, any $X_i \in \sS$ that induces the conditional dependency as described above implies an edge going from $i$ to $j$. The number of such predecessors is equivalent to the in-degree of node $j$. The test is repeated on every node in the ordering, practically with $\mathsf{Chi}$-$\mathsf{square}$ independence tests on $5,000$ samples  at a significance level of $0.01$ in this case.

\input{inputs/algo2}

Suppose the maximum in-degree of $\rmG$ is $\mathrm{deg}^{-}_{\max}$ (i.e., every node has at most $\mathrm{deg}^{-}_{\max}$ incoming edges), any node $j$ with the estimated in-degree from $\pi$ large than $\mathrm{deg}^{-}_{\max}$ is thus potentially invalid in that a parent $j$ might be placed later and/or a descendant of $j$ might be placed earlier in the ordering. In other words, $j$ is positioned incorrectly relative to their parents and descendants.  We note that the test is for diagnosis purposes only and the conclusion is not definitive unless faithfulness is strongly assumed along with an accurate knowledge of sparsity bounds. However, since sparsity is fairly common in practice, if this occurs to most or all nodes in $\pi$, which would effectively imply a fully connected DAG, there is a high chance of structural mismatch.

\paragraph{Empirical Analysis.} This insight motivates us to study the correlation between the proportion of nodes with in-degree of higher than $\mathrm{deg}^{-}_{\max}$ with topological divergence metric $\mathsf{D}_{top} (\downarrow)$, which quantifies the true error of $\pi$. If the true maximal degree is $\mathrm{deg}^{-}_{\max}$, nodes with in-degree of more than $\mathrm{deg}^{-}_{\max}$ are more likely to be incorrect. Therefore, we expect a positive correlation where higher proportion of invalid nodes corresponds to higher $\mathsf{D}_{top}$. 

We generate $1,000$ random ER graphs with number of nodes $d \sim U(5, 30)$ and expected degree $kd$ where $k \in U(2,4)$. We consider random permutations of nodes as candidate estimations of causal order.  Figure \ref{fig:corr} illustrates the correlation between $\mathsf{D}_{top}$ and the in-degree distributions estimated from the topological orders given the knowledge of $\mathrm{deg}^{-}_{\max} = 3$. We here normalize $\mathsf{D}_{top}$ over the maximal number of edges in DAGs of ${d \choose 2}$ to remove the effect of graph size. A positive correlation can be observed in the illustration. We confirm this observation with Kernel-based conditional independence (KCI) tests \citep{zhang2011kernel} (with Polynomial kernel), which verify the dependency between $\mathsf{D}_{top}$ and the in-degree distributions, conditioned on number of nodes and edges in $\rmG$. Figure \ref{fig:pvalue} reports the $p$-values of the tests at different hypotheses of $\mathrm{deg}^{-}_{\max}$ and demonstrates a strong evidence for the conditional dependency between these two quantities. This indicates that the in-degree distribution induced by a candidate ordering can serve as an indicator for the validity of the ordering w.r.t the true DAG $\rmG$.

%% file: inputs/algo2.tex
\begin{algorithm}[t!] 
    \caption{Estimation of In-degree Distribution via Conditional Independence Test}
    \label{algo:markov}
\begin{algorithmic}
\State \textbf{Input:} Data matrix $X \in [n]^{N \times d}$ and estimated ordering $\pi$. 

\State \textbf{Output:} In-degrees of non-root nodes;

\State Initialize \texttt{InDegree} $\gets \boldsymbol{0}_{d-1} $; \Comment{Zero vector of dimension $d-1$}
\For{$k = 2, \cdots,d$} 

    \State $X_j \gets X_{v \mid v \in [d], \pi_v = k}$;
    
    \State $ \sS \gets \{ X_i : i \in [d], \pi_i < k \} $; \Comment{Set of node $j$'s predecessors}
    
     \For{$ X_i \in \sS $}
        \State $ \sS_{\text{cond}} \gets \sS \setminus \{X_i\} $; \Comment{Conditioning set without $ X_i $}
        
        \If{$ X_i \not\indep X_j \mid \sS_{\text{cond}} $}
            \State \texttt{InDegree[j]} $\gets \texttt{InDegree[j]} + 1$; \Comment{Update in-degree of node $j$}
        \EndIf
    \EndFor
    \
\EndFor
\end{algorithmic}
\end{algorithm}

%% file: inputs/Conclusion.tex
\section{Limitations and Conclusion}

In this work, we have explored the connection of discrete score matching to causal discovery and contributed to the score matching literature a novel identifiability result to infer causal orders from observational discrete data. Theoretically, Theorem \ref{theorem:main} is as well applicable to continuous or mixed data. Nonetheless, a direct application in these domains is inherently challenging due to the statistical bottleneck of estimating singleton conditional densities. We are unaware of any tractable estimation approach for continuous variables with formal guarantees, thus have focused only on the discrete domain, which itself remained underexplored.  

In terms of the order search \underline{alone}, the time complexity is linear in the number of nodes. The dominant factor is the training time of the continuous-time diffusion model for estimating the discrete score functions. The algorithm involves recursively estimating the score function from the data where the identified leaf variable is removed. Consequently, the model must be retrained at every iteration, which unfortunately increases the training time. Future improvements in model design, through possibly adaptive masking strategies, may enable single-model approximation of the discrete scores with various missing data patterns. This could effectively reduce the order search complexity strictly to $\gO(d)$. 

Finally, as in any identifiability theory, our condition necessitates prior knowledge about the data-generating process, which may be lacking in real-world scenarios. We leave the exploration of these open questions to future research.


%% file: inputs/Proof.tex
\section{Proofs}

\input{lib/proof-thm1}

\input{lib/proof-prop1}

%% file: lib/proof-thm1.tex
\subsection{Proof for Theorem 
\ref{theorem:main}}\label{apdx:proof-thm}

We first need to prove that under Condition \ref{ass:strict}, for every non-leaf node $i$ in the SCM, conditioning on the full Markov blanket, must strictly reduce the randomness of $X_i$ compared with conditioning only on its parents. Thus, only leaf nodes have no extra Markov blanket information beyond their parents, since $\Mb{i}=\Pa{i}$ if $i$ is a leaf node and hence $\phi(X_i\mid X_{\Mb{i}})=\phi(X_i\mid X_{\Pa{i}})$.

\input{lib/proof-lm1}

\textbf{Theorem \ref{theorem:main}.}
\textit{Let $x \in \gX$ be a discrete random vector defined via an SCM (\ref{eq:scm}), and let $\rvr_i(x_{-i}) := p(X_i \vert x_{-i})$ be the reciprocal discrete score function for every node $i \in [d]$. If there exists a randomness measure $\phi$ satisfying the non-decreasing randomness property w.r.t the true graph $\rmG$, then
    $X_j \text{ is a leaf node } \Leftrightarrow j \in \argmax_{i\in [d]} \mathbb{E}_{X_{-i}} \Big[ \phi\big( \rvr_i(x_{-i}) \big) \Big]$.
}

\begin{proof}



For ease of notation, let $\rvp_i(x_{\Pa{i}}) := p(X_i \vert x_{\Pa{i}})$ for any $i \in [d]$. 

Consider a leaf node $l$, we have $p(x_l \vert x_{-l}) = p(x_l \vert x_{\Pa{l}})$. Hence 
\begin{align}
\mathbb{E}_{X_{-l}} \Big[ \phi(\rvr_l) \Big] =   \mathbb{E}_{X_{-l}} \Big[ \phi(\rvp_l) \Big]  = \phi(X_l \vert X_{\Pa{l}}).
\end{align}

If $i$ is a non-leaf node, we have 
\begin{align}
    \mathbb{E}_{X_{-i}} \Big[ \phi(\rvr_i) \Big] =   \phi(X_i \vert X_{-i}) = \phi(X_i \vert X_{\Mb{i}}).
\end{align}

\paragraph{\textbf{Proof for the $``\Rightarrow"$ direction.}} 

For any non-empty subset of nodes $\rmS \subseteq \rmV$, let $\rmG[\rmS]$ be the induced subgraph on $\rmS$. We first show that by Condition \ref{ass:non-dec}, every subgraph $\rmG[\rmS]$ yields an order by the randomness criterion $\phi$ that always terminates at a leaf node.

Let $\rmM := \argmax_{i \in \rmS} \phi(X_i \vert X_{\Pa{i}})$ be the set of $\phi$-maximizers in $\rmS$. Take any $i \in \rmM$. If $(i,j)$ is an edge in $\rmG[\rmS]$, then by monotonicity, $\phi(X_i \vert X_{\Pa{i}}) \le \phi(X_j \vert X_{\Pa{j}})$, hence $j \in \rmM$. Therefore all children of a maximizer are also maximizers i.e., $\rmM$ is upward closed in $\rmG[\rmS]$.

Since $\rmG[\rmS]$ is a finite DAG, the induced subgraph $\rmG[\rmM]$ has at least one leaf node. Also $\rmM$ has no edges to $\rmS \backslash \rmM$, thus a leaf of $\rmG[\rmM]$ is also a leaf of $\rmG[\rmS]$. Hence, there exists a leaf node $l \in \rmS$ such that $\phi(X_l \vert X_{\Pa{l}}) = \max_{i \in \rmS} \phi(X_i \vert X_{\Pa{i}})$, which proves our desired property. 

Applying the conditional information inequality and Condition \ref{ass:non-dec} respectively, one has that
\begin{align}
 \phi(X_i \vert X_{\Mb{i}}) \le \phi(X_i \vert X_{\Pa{i}}) \le \phi(X_l \vert X_{\Pa{l}}).    
\end{align}

By Markov property, $\Mb{i}$ is the minimal set of nodes that renders $X_i$ independent from the other variables. Also by Lemma \ref{lm:strict}, the first inequality is therefore strict. 

We conclude that $\mathbb{E}_{X_{-l}} \Big[ \phi(\rvr_l) \Big] > \mathbb{E}_{X_{-i}} \Big[ \phi(\rvr_i) \Big], \forall i \ne l$.

\paragraph{\textbf{Proof for the $``\Leftarrow"$ direction.}}

Suppose there exists a non-leaf node $i$ such that $i = \argmax_{i \in [d]} \mathbb{E}_{X_{-i}} \phi(\rvr_i)$. It follows that for every leaf node $l$,  
\begin{align*}
   \mathbb{E}_{X_{-i}} \phi(\rvr_i) = \phi(X_i \vert X_{\Mb{i}}) > \mathbb{E}_{X_{-l}} \phi(\rvr_l) = \phi(X_l \vert X_{\Pa{l}}), \quad \forall l \ne i, \Ch{l} = \emptyset.
\end{align*}

Since $\rmG$ is a finite DAG, there exists one leaf node $l$ that is a descendant of $i$ via a directed path. As $\phi$ satisfies the non-decreasing randomness condition, one has that $\phi(X_l \vert X_{\Pa{l}}) \ge \phi(X_i \vert X_{\Pa{i}})$. This leads to $\phi(X_i \vert X_{\Mb{i}}) > \phi(X_i \vert X_{\Pa{i}})$, which contradicts the conditional information inequality given by Condition \ref{ass:strict}. 

Therefore, $i$ cannot have descendants, rendering  
 $\phi(X_i \vert X_{\Mb{i}}) = \phi(X_i \vert X_{\Pa{i}})$, hence $i$ must be a leaf node. 
\end{proof}

%% file: lib/proof-lm1.tex
\begin{lemma}[Strict Markov blanket gain]\label{lm:strict}
Let $x \in \gX$ be a discrete random vector defined via an SCM (\ref{eq:scm}). Suppose Condition \ref{ass:strict} holds w.r.t every non-leaf variable $X_i$ in the SCM. Then for any measure of randomness $\phi$ defined in Eq. (\ref{def:rand-form}), 
$$
\phi(X_i\mid X_{\Pa{i}}) > \phi(X_i\mid X_{\Mb{i}}). 
$$
\end{lemma}

\begin{proof}

By the law of total probability, 
\begin{align}
    p(X_i \mid x_{\Pa{i}}) = \sum_{x_{b_i}} p(x_{b_i} \mid x_{\Pa{i}}) p(X_i \mid x_{\Pa{i}}, x_{b_i})
\end{align}
for each node $i$ and parent configuration $x_{\Pa{i}}$. By concavity of $\phi$, 
\begin{align}
    \phi(X_i \mid x_{\Pa{i}}) \ge \sum_{x_{b_i}} p(x_{b_i} \mid x_{\Pa{i}}) \phi(X_i \mid x_{\Pa{i}}, x_{b_i}). 
\end{align}

For every $x_{\Pa{i}} \in \gA_i$, by Condition \ref{ass:strict}, there exist $x_{b_i} \neq x'_{b_i}$ with
$p(x_{b_i}\mid x_{\Pa{i}}), p(x'_{b_i} \mid x_{\Pa{i}})>0$ and $p(X_i \mid x_{\Pa{i}}, x_{b_i})\neq p(X_i \mid x_{\Pa{i}}, x'_{b_i})$. Thus the above mixture is non-trivial.
Since $\phi$ is strictly concave, Jensen's inequality is strict, hence for all $x_{\Pa{i}} \in \gA_i$
\begin{align}
    \phi(X_i \mid x_{\Pa{i}}) > \sum_{x_{b_i}} p(x_{b_i} \mid x_{\Pa{i}}) \phi(X_i \mid x_{\Pa{i}}, x_{b_i}).
\end{align}

Since $\Pr(X_{\Pa{i}}\in \gA_i) > 0$, taking expectation yields
\begin{align}
    \E_{X_{\Pa{i}}} \Big[ \phi(X_i \mid x_{\Pa{i}}) \Big] > \sum_{x_{\Pa{i}}} \sum_{x_{b_i}} p(x_{\Pa{i}}) p(x_{b_i} \mid x_{\Pa{i}}) \phi(X_i \mid x_{\Pa{i}}, x_{b_i}),
\end{align}
or equivalently 
\begin{align}
\phi( X_i \mid X_{\Pa{i}}) > \phi( X_i \mid X_{\Mb{i}}).
\end{align}
This completes the proof.
\end{proof}

%% file: lib/proof-prop1.tex
\subsection{Proof for Proposition \ref{prop:main}}\label{apdx:proof-prop}

\textbf{Proposition \ref{prop:main}.}
\textit{
Let $x \in \gX$ be a discrete random vector defined via a symmetric Dirichlet SCM (\ref{eq:scm-dir}). For sufficiently large $\alpha_0$, Condition \ref{ass:non-dec} holds w.r.t the Shannon entropy function $H(\cdot)$ if
\begin{align}
\forall i \in[d], \ \forall  j \in \Ch{i}: \quad \log \left( \frac{n_j}{n_i} \right) 
+ \alpha^{-1}_{0} \left[ K_i c(n_i) - K_j c(n_j) \right] \ge 0,
\end{align}
 with $ c(n) := \frac{3}{2}n + \frac{1}{2} $.
}

\begin{proof}

Let $s_i := \alpha_0 \slash K_i$ be the total concentration. For a symmetric Dirichlet with total row concentration $s_i$ on a $n_i$-simplex, the expected entropy is given as 
\begin{align}
    H(X_i \mid X_{\Pa{i}})&  := \E_{X_{\Pa{i}}} \left[ H(X_i \mid x_{\Pa{i}}) \right], \nonumber \\ 
    & = \psi(s_i + 1) -  \psi \left( \frac{s_i}{n_i} + 1 \right),
\end{align}
where $\psi$ is the digamma function.

For large $\alpha_0$, using the approximation $\psi(x) \approx \log x - \frac{1}{2x}$ gives 
\begin{align}
    & H(X_i \mid X_{\Pa{i}}) \nonumber \\ 
    & \approx \log \left( n_i \cdot \frac{\alpha_0 + K_i}{\alpha_0 + n_i K_i} \right) - \frac{K_i}{2(\alpha_0 + K_i)} + \frac{n_i K_i}{2 (\alpha_0 + n_i K_i)}.
\end{align}

Given two variables $X_i$ and $X_j$, we define the entropy gap:
\begin{align}
\Delta H_{i,j} := H(X_j \mid X_{\Pa{j}}) - H(X_i \mid X_{\Pa{i}}).
\end{align}

Using the approximation above, we have that
\begin{align}
\Delta H_{i,j} 
&\approx \log \left( \frac{n_j}{n_i} \cdot \frac{\alpha_0 + K_j}{\alpha_0 + K_i} \cdot \frac{\alpha_0 + n_i K_i}{\alpha_0 + n_j K_j} \right) \notag \\
&+ \left( \frac{K_i}{2(\alpha_0 + K_i)} - \frac{n_i K_i}{2(\alpha_0 + n_i K_i)} \right) \nonumber \\
&- \left( \frac{K_j}{2(\alpha_0 + K_j)} - \frac{n_j K_j}{2(\alpha_0 + n_j K_j)} \right).
\end{align}

In the large-$ \alpha_0 $ regime, we expand all terms to leading order in $ 1/\alpha_0 $. Using:
\[
\log \left(1 + \frac{x}{\alpha_0}\right) \approx \frac{x}{\alpha_0}, \quad 
\frac{1}{\alpha_0 + x} \approx \frac{1}{\alpha_0} - \frac{x}{\alpha_0^2},
\]
we approximate:
\begin{align}
\Delta H_{i,j} &\approx \log \left( \frac{n_j}{n_i} \right)
+ \frac{1}{\alpha_0} \left[ (1 + n_i)K_i - (1 + n_j)K_j \right] \nonumber \\ 
& + \frac{1}{2\alpha_0} \left[ K_i(n_i - 1) - K_j(n_j - 1) \right]  \\
&= \log \left( \frac{n_j}{n_i} \right) 
+ \frac{1}{\alpha_0} \Bigg[ K_i \left( n_i + \frac{1}{2}(n_i - 1) + 1 \right) \nonumber \\
& - K_j \left( n_j + \frac{1}{2}(n_j - 1) + 1 \right) \Bigg] .
\end{align}

Let us define $c(n) := \frac{3}{2}n + \frac{1}{2}$ so that:
\begin{align}
\Delta H_{i,j} \approx \log \left( \frac{n_j}{n_i} \right) 
+ \frac{1}{\alpha_0} \left[ K_i c(n_i) - K_j c(n_j) \right].
\end{align}

Condition \ref{ass:non-dec} holds w.r.t the Shannon entropy function $H(\cdot)$ iff 
\begin{align}
  \forall i \in [d], \quad \forall j \in \Ch{i}: \quad  H(X_i \mid X_{\Pa{i}}) \le H(X_j \mid X_{\Pa{j}}).   
\end{align}

It is easy to see that the inequality is equivalent to $\Delta H_{i,j} \ge 0$, or
\begin{align}
 \log \left( \frac{n_j}{n_i} \right) 
+ \frac{1}{\alpha_0} \left[ K_i c(n_i) - K_j c(n_j) \right] \ge 0.
\end{align}

\end{proof}

%% file: inputs/APDX-RWork.tex
\section{Related Work}\label{sec:rwork}
Causal discovery algorithms can be broadly categorized as either constraint-based methods, such as \texttt{PC} \citep{spirtes1991algorithm} and \texttt{FCI} \citep{spirtes2000causation} which detect edge existence and direction via conditional independence tests, or score-based methods that optimize for DAGs based on a given objective function
\citep{ott2003finding,chickering2002optimal,teyssier2012ordering,cussens2017polyhedral}. Research on continuous data particularly enjoys remarkable progress over the years, driven by the development of non-convex characterization of the acyclicity constraints. This gives rise to a family of scalable DAG learning frameworks via continuous optimization programs, notably in \citet{lachapelle2019gradient,zheng2020learning,yu2019dag,bello2022dagma}. 

Constraint-based causal discovery can be extended to discrete data with $\mathsf{G}$-tests \citep{quine1985efficiencies} or ${\chi}^2$ tests \citep{cochran1952chi2}. Score-based methods, such as \texttt{GES} \citep{chickering2002optimal,teyssier2012ordering}, can be applied on multinomial Bayesian networks with BIC \citep{schwarz1978estimating} or BDeu \citep{heckerman1995learning} scoring functions. However, it is well-known that under these approaches, causal graphs are only identifiable from observational data up to the Markov equivalence class. Several identifiability results have thus been proposed, under specific assumptions, for nominal/categorical data \citep{peters2010identifying,liu2016causal,cai2018causal,kocaoglu2017entropic,compton2022entropic,qiao2021learning}, ordinal data \citep{luo2021learning,ni2022ordinal}, or mixed data \citep{tsagris2018constraint,sedgewick2019mixed,wenjuan2018mixed}. These existing methods are mainly designed for identifying bivariate causal direction and algorithmically, they typically resort to constraint-based or score-and-search algorithms. 


\paragraph{\textbf{Ordering-based Causal Discovery.}}
Ordering-based causal discovery exploits the fact that the space of variable orderings is substantially smaller than the space of DAGs, and searching over topological orders avoids the need to explicitly impose an acyclicity constraint \citep{teyssier2012ordering,buhlmann2014cam}. Research in ordering-based causal discovery recently takes off with the use of score matching. These methods  assume observational data arise from a continuous additive noise model. Rather than searching over the full ordering space or relying primarily on sparsity, they recover the ordering based on a leaf identification principle: one sequentially identifies a leaf node, removes it from the graph, and repeats the procedure, with the final causal order obtained by reversing the leaf-removal sequence ~\citep{ghoshal2018learning,chen2019causal,rolland2022score,sanchez2022diffusion,montagna2023shortcuts,montagna2023causal,montagna2023scalable,xuordering}. This family of approaches has been shown to improve robustness to assumption violations \citep{montagna2024assumption}, scalability to high-dimensional graphs \citep{montagna2023scalable}, and provides finite-sample complexity guarantees \citep{zhu2024sample}.

\paragraph{\textbf{Score Matching.}} 
Score matching is a family of parameter learning methods alternative to the maximum likelihood principle. The objective entails matching two log probability density functions by their first-order derivatives using the Fisher divergence. First introduced in \citep{hyvarinen2005estimation}, score matching obviates the intractability of the normalizing partition functions as well as the ground-truth data score, while yielding a consistent estimate. Further developments in score estimation include kernel-based estimators \citep{li2017gradient}, denoising score matching \citep{vincent2011connection}, slice score matching \citep{song2020sliced}, denoising likelihood score matching \citep{chao2022denoising}, and score-based generative modelling \citep{song2019generative}. The score function therein can be well approximated by a deep neural network and learned by minimizing the empirical Fisher divergence. 

While representing a probability distribution by the score of its density has proven effective for continuous data, the notion of gradient is not defined for discrete modalities, rendering score matching inapplicable. To this end, a popular surrogacy of the typical score function is proposed as what is known as the \textit{concrete score} \citep{meng2022concrete}, which is the ratio of two marginal probabilities for different state-value pairs $\frac{p(y)}{p(x)}$. Analogous to the score function $\nabla \log p(x)$, this ratio arises in the reverse process for discrete diffusion models, where the evolution of discrete variables is described through a continuous-time Markov chain \citep{anderson2012continuous,campbell2022continuous,sun2022score,loudiscrete}, leading to a natural realization of the score function in discrete domains. 

Another direction focuses on categorical score estimation by matching marginal probabilities for each dimension. This approach, called \textbf{ratio matching}, is initially proposed by \citet{hyvarinen2007some} for binary data, where it also preserves consistency (under some regularity conditions) and bypasses the computation of normalizing constant. Extensions to general discrete data are developed in \citet{lyu2012interpretation} and \citet{sun2022score}, leading to the framework of generalized score matching described in Section \ref{sec:gsm}.